\documentclass{article}

\usepackage[final]{nips_2017}

\usepackage[utf8]{inputenc} % allow utf-8 input
\usepackage[T1]{fontenc}    % use 8-bit T1 fonts
\usepackage{hyperref}       % hyperlinks
\usepackage{url}            % simple URL typesetting
\usepackage{booktabs}       % professional-quality tables
\usepackage{amsfonts}       % blackboard math symbols
\usepackage{nicefrac}       % compact symbols for 1/2, etc.
\usepackage{microtype}      % microtypography

\usepackage{amsmath}								
\usepackage{amssymb}							
\usepackage{graphicx}
\usepackage{subcaption}
\DeclareMathOperator*{\argmin}{arg\,min}
\usepackage{color, colortbl}
\definecolor{Gray}{gray}{0.9}
\usepackage{tabularx}
\usepackage{xspace}

\title{Disentangled Representations for Manipulation of Sentiment in Text}

\author{
  Maria Larsson \\
  Sigma Embedded Engineering, Sweden  \\
   \texttt{maria.larsson@sigma.se} \\
  \And 
  Amanda Nilsson \\
  Zenuity, Sweden \\
   \texttt{amanda.nilsson@zenuity.com} \\
  \And Mikael K\r{a}gebäck \\
  Chalmers University of Technology, Sweden  \\
  \texttt{kageback@chalmers.se} \\
}

\begin{document}
% \nipsfinalcopy is no longer used

\maketitle

\begin{abstract}

The ability to change arbitrary aspects of a text while leaving the core message intact could have a strong impact in fields like marketing and politics by enabling e.g. automatic optimization of message impact and personalized language adapted to the receiver's profile. In this paper we take a first step towards such a system by presenting an algorithm that can manipulate the sentiment of a text while preserving its semantics using disentangled representations. Validation is performed by examining trajectories in embedding space and analyzing transformed sentences for semantic preservation while expression of desired sentiment shift.

\end{abstract}

\section{Introduction}
%Motivation
As we live in an increasingly digitized society, algorithms for text analysis and generation can be used for a variety of purposes and may greatly relieve manual work. A system for robust manipulation of global text properties, e.g. sentiment, is one such algorithm that could potentially change how we work with text and open up new possibilities. Though the main purpose of a text might be to communicate a concrete message there are an infinite number of ways the message can be phrased, each with an individual set of global properties connected to it. 
In this paper we focus on the sentiment aspect and note that robust control over the sentiment would open up a range of new possibilities, like AB testing of different instantiations of a message with respect to some desired measure, and personalized communication automatically adapted to the receiver's profile. Further, the ability of generating new sentences with transformed sentiment could also be useful in data augmentation when the available data is scarce.

%\subsection{Related work}
% controllable text generation
Recent work in text generation \citep{hu2017controllable,radford2017learning} has shown that it is possible to generate random sentences where the sentiment can be chosen as an input parameter. This line of research has some similarities to the problem we are addressing in this paper but with the key difference that while they generate new random sentences we aim to transform existing sentences. This makes the problem more difficult but also more applicable to real world applications as shown by the recent work of \cite{mueller2017sequence}. 
% visual manifold traversal

In the visual domain there has been a range of work lately that aims to transform the input image to fit different aspects, e.g. to look like a painting \citep{gatys2015neural}. The method presented by \cite{DeepManifold2016} transforms an image to a deep feature space using a convolutional neural network (CNN). This space is then traversed towards the target features. A new image is subsequently reconstructed from the deep feature representation but where some aspect has been changed from the original image. In their experiments they show that this can be used to transform a smiling portrait into an angry one and make one individual look more like someone else without changing clothing or background. The method we present in this paper is loosely based on their model, however, with significant changes due to the discrete nature of language. 

%Main contributions
The main contributions of this work include: 
(1) an algorithm that can automatically transform the sentiment of a text while leaving the semantic content largely intact, and (2) preliminary qualitative analysis of the performance with regard to (a) resulting sentiment, (b) semantic stability and (c) acceptability of the transformed text.

%\section{Background}

%
%---------------------------------------
\section{Maximum mean discrepancy}
\label{sec:mmd}
%---------------------------------------
%
The maximum mean discrepancy (MMD) \citep{gretton2012kernel} is a test statistic used to determine whether two distributions are the same. 
%This statistic is useful when, for example, determining whether measurements from two setups of the same experiment may be analyzed jointly. Another application is to use the statistic for distinguishing sick people from healthy people, when analyzing tissue samples \citep{gretton2012kernel}.
Given two distributions, $\mathcal{P}_\text{source}$ and $\mathcal{P}_\text{target}$, the objective of the MMD is to find a smooth function which is large for samples from $\mathcal{P}_\text{source}$ and small for samples from $\mathcal{P}_\text{target}$. Given such a function the MMD is the difference between the mean function values for the two sets of samples, which can be empirically estimated as $\text{MMD}(\mathcal{F}, X, Y) =$
%. \newcite{gretton2012kernel} presents an empirical estimate of the $\text{MMD}(\mathcal{F}, X, Y) =$
%\begin{equation}
%\begin{split}
%    &\text{MMD}(\mathcal{F}, X, Y) = \\
%    &\sup\limits_{f \in \mathcal{F}}( \frac{1}{m}\sum\limits_{i = 1}^m f(x_i) - \frac{1}{n} \sum\limits_{i = 1}^n f(y_i))
%\end{split}
    %\label{eq:MMD}
%\end{equation}
\begin{equation}
    \sup\nolimits_{f \in \mathcal{F}}\left( \sum\nolimits_{i = 1}^m \frac{f(x_i)}{m} -  \sum\nolimits_{i = 1}^n \frac{f(y_i)}{n}\right)
    \label{eq:MMD}
\end{equation}
where $X = [x_1,x_2, \dots, x_m]$ are samples drawn from the source distribution $\mathcal{P}_\text{source}$ and $Y = [y_1, y_2, \dots, y_n]$ are samples drawn from the target distribution $\mathcal{P}_\text{target}$. The function $f$ belongs to a class, $\mathcal{F}$, of smooth functions and should be chosen as to maximize the difference between the mean values of $f$ applied to $X$ and $Y$. In both \citep{gretton2012kernel} and \citep{DeepManifold2016}, $\mathcal{F}$ is a reproducing kernel Hilbert space allowing comparison of multi-dimensional feature vectors. The function $f^*$ attaining the supremum in equation \eqref{eq:MMD} can be empirically estimated as
\begin{equation}
    f^*(z) = \frac{1}{m}\sum\limits_{i = 1}^m k(x_i,z) - \frac{1}{n} \sum\limits_{i = 1}^n k(y_i,z),
    \label{eq:witness}
\end{equation}
where $k(x,x')$ is a kernel function. The method presented by \cite{DeepManifold2016} uses a Gaussian kernel function
$
    k(x,x') = e^{-\frac{1}{2\sigma}|x - x'|^2}
$
with $\sigma$ being the kernel bandwidth.
%

%%%%%%%%%%%%%%%%%%%%%%%%%%%%%%%%
\section{Model}
%%%%%%%%%%%%%%%%%%%%%%%%%%%%%%%%
The problem we are addressing can be split into three different subtasks. The first task is representing sentences in a continuous space. The second task is exploiting the sentence representation and traversing the manifold in such a way that the sentiment changes. The third task is generating a sentence from the representation space. Our model uses a CNN for sentence encoding. The encoded vectors are subsequently traversed using the MMD statistic and finally decoded using a recurrent neural network (RNN).

% =====================================================================
\subsection{Encoding sentences }
% =====================================================================
A sentence is represented as a matrix where the rows correspond to the, 300-dimensional,  \emph{word2vec} \citep{mikolov2013distributed} word embeddings for each word in the sentence. This matrix is given as input to a CNN, trained for binary sentiment classification. The network consists of one convolutional layer, one max-pooling layer and finally one fully connected feed forward layer. The filter heights for the convolutional layer are $1,2,3$ and $4$, and the filter width is 300. 75 filters per size results in a total of 300 filters. The pooling layer therefore outputs a 300-dimensional feature vector, denoted $\mathbf{z}$. This feature vector is extracted from the CNN, along with the predicted label, and used as the encoding of the input sentence.

\begin{figure}
	\begin{minipage}{.48\textwidth}
    \centering
    \includegraphics[width=1.0\linewidth]{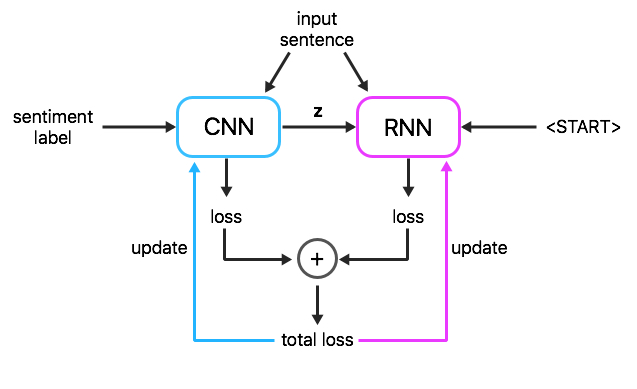}
    \caption[Training scheme for the CNN and RNN]{During training, the CNN and RNN are updated using the unweighted sum of the loss for sentiment classification and for text generation.}
    \label{fig:training}
\end{minipage}%
~
\begin{minipage}{.49\textwidth}
   	\centering
   	\includegraphics[width=0.5\linewidth, trim={3cm 2cm 1cm 1cm},clip]{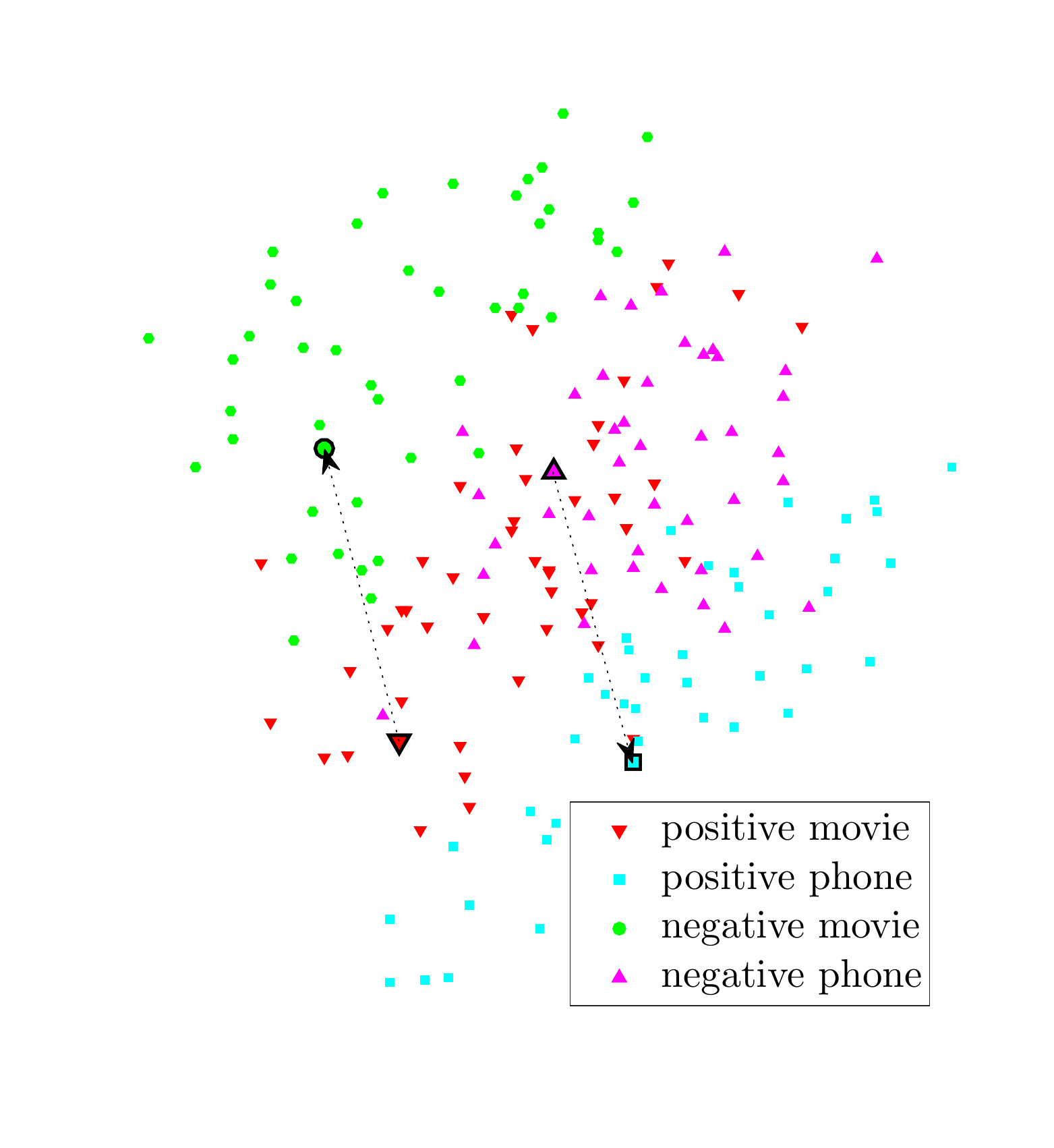}
   	\caption{Different icons distinguish feature vectors by sentiment and topic. Bold faced points are examples of original and traversed vectors. }
   	\label{fig:movie_phone_traversal}
\end{minipage}
\end{figure}

In addition to classifying sentiment, the CNN needs to encode information about the topic and semantics of the sentence. Therefore, it is trained together with the RNN. Initially, the sentiment classification task is disregarded and the joint networks are trained for encoding and decoding unlabeled sentences. The loss for this task is measured by calculating the cross-entropy error between the predicted word, $\hat{w}$, at position $t$, in the generated sentence and the actual word, $w$, at the same position from the original sentence. After this initial training phase, the CNN is trained on binary sentiment classification. The classification loss is calculated as the cross-entropy error between the predicted label and the true label for each sentence. This loss is added to the text generation loss, producing a total loss which is used to update the weights in both networks. A schematic of the training procedure is illustrated in figure \ref{fig:training}. 

% =====================================================================
\subsection{Traversal of the representation space}
% ===================================================================== 
Since the CNN is trained on binary sentiment classification, two separable distributions of feature vectors are generated. The MMD statistic can be used to traverse a vector originating from one of these distributions to the other. The result of the traversal is a vector that resembles the encoding of a sentence with the opposite sentiment.

When moving the feature vector $\mathbf{z}$ by minimizing equation \eqref{eq:witness}, the semantics of the original sentence may be lost if $\mathbf{z}$ is moved too far along the manifold. To control how far $\mathbf{z}$ is moved from its original position a \emph{budget of change} \citep{DeepManifold2016}, $\lambda$, is used. A source and a target set of sentence representations are created. The source set, $\mathbf{z}^s$, contains feature vectors for sentences with the same sentiment as $\mathbf{z}$ and the target set, $\mathbf{z}^t$, contains feature vectors for sentences with the opposite sentiment. From these sets and the original vector, a matrix $\mathbf{V} = [\mathbf{z}^t_1, \cdots, \mathbf{z}^t_n, \mathbf{z}^s_1, \cdots, \mathbf{z}^s_m, \mathbf{z}]$ is formed. The traversed feature vector can then be expressed as $\mathbf{z}^* = \mathbf{z} + \mathbf{V}\mathbf{\delta}$, where $\mathbf{\delta}$ is a coefficient vector. Equation \eqref{eq:witness} can now be written as 
$ f^*(\mathbf{z} + \mathbf{V\delta}) = \frac{1}{m}\sum\limits_{i = 1}^m k(\mathbf{z}^s_i,\mathbf{z} + \mathbf{V\delta}) - \frac{1}{n} \sum\limits_{i = 1}^n k(\mathbf{z}^t_i,\mathbf{z} + \mathbf{V\delta}),$
%\begin{equation}
%\begin{split}
%    f^*(\mathbf{z} + \mathbf{V\delta}) = &\frac{1}{m}\sum\limits_{i = 1}^m k(\mathbf{z}^s_i,\mathbf{z} + \mathbf{V\delta}) - \\
%    &\frac{1}{n} \sum\limits_{i = 1}^n k(\mathbf{z}^t_i,\mathbf{z} + \mathbf{V\delta}),
%\end{split}
%    \label{eq:f_star}
%\end{equation}
where $ \mathbf{\delta} = \argmin_{\mathbf{\delta}} f^*(\mathbf{z} + \mathbf{V\delta}) + \lambda\|\mathbf{V\delta} \|^2$, $\lambda \in \mathbb{R}$. The minimization over $\delta$ uses the BFGS algorithm \citep{battiti1990optimization} and is constrained by the budget of change, enforced in the last term.

%\begin{equation}
%\label{eq:budget}
%\begin{split}
%    \mathbf{\delta} &= \argmin_{\mathbf{\delta}} f^*(\mathbf{z} + \mathbf{V\delta}) + \lambda\|\mathbf{V\delta} \|^2 , \\
%    &\qquad \lambda \in \mathbb{R}.
%\end{split}
%\end{equation}
%The minimization over $\delta$ uses the BFGS algorithm \citep{battiti1990optimization}. The optimization of \eqref{eq:budget} is constrained by the budget of change, enforced in the last term. 

% =====================================================================
\subsection{Decoding sentences}
% =====================================================================
The traversed feature vector $\mathbf{z}^*$ is given as input to an RNN trained for generating text. In addition to $\mathbf{z}^*$, the RNN receives a start-of-sentence token as input in the first time step. For each time step, the RNN outputs the most probable word and gives this word as input to the next time step. When the most probable word is an end-of-sentence token, the generation of words is terminated. The RNN consists of a single layer GRU cell, with a state size of 300. The weight matrix for the input, $\mathbf{W}_x$, consists of the 300-dimensional \emph{word2vec} word embeddings for the words in the vocabulary.

\section{Experiments and results}
The initial encoding and decoding training uses the large movie review dataset v1.0 \citep{maas-EtAl:2011:ACL-HLT2011} disregarding the label. The networks are then trained on three sentiment labelled data sets. The first set is the movie review sentence polarity data set v1.0\footnote{https://www.cs.cornell.edu/people/pabo/movie-review-data/} \citep{Pang+Lee:05a} which consists of 10 662 labelled movie-review sentences from www.rottentomatoes.com. The second set contains 500 reviews for cell phones and accessories from Amazon, 500 reviews for restaurants from Yelp and 500 movie reviews from IMDB\footnote{https://archive.ics.uci.edu/ml/machine-learning-databases/00331/} \citep{DBLP:conf/kdd/KotziasDFS15}. These two sets have equal amounts of positive and negative sentences. The third set is a subset of 923 positive and 1320 negative sentences from a data set\footnote{https://github.com/oscartackstrom/sentence-sentiment-data} containing product reviews from various online sources \citep{Tackstrom}. The three data sets are randomly divided 90\%-10\% into a training and a test set. The training set is used for updating the weights of the networks during training and is divided into batches of 64 sentences. The test set is used for evaluating the accuracy of the networks periodically during training.

%------------------------------------------
\subsection{Preserving semantics}
%------------------------------------------
In order to evaluate whether the encodings from the CNN contain information about sentiment and semantics, feature vectors for the sentences with different sentiments and topics are visualized. These visualizations also serve as an aid for assessing whether the content in a sentence is preserved in the traversal. The feature vectors are reduced from 300 to 2 dimensions using principal component analysis (PCA) and the visualizations are made using the first two principal components. 

The choice of topics was sentences containing either the word \emph{phone} or \emph{movie}, because such sentences would likely have little correlation in contrast to, for example, sentences containing either \emph{comedy} or \emph{drama}. Negative sentences containing the word movie and positive sentences containing the word phone were traversed. The optimization of the MMD was set up with 90 positive examples and 90 negative examples for the source and target sets, and $\lambda = 7$e$-5$. The examples consisted of an equal amount of sentences containing the word movie and sentences containing the word phone. The topics of the sentences were not used for the traversal but needed when visualizing the results.

The results are shown in figure \ref{fig:movie_phone_traversal}. It is seen that a vector representing a positive sentence containing movie is moved so that the resulting vector lies within the cluster of negative sentences containing movie. In the same way, a vector representing a negative sentence containing phone is moved so that the resulting vector lies within the cluster of positive sentences containing phone. This behaviour suggests that the context and semantics may be preserved during the traversal.

Since the manifold traversal is made using two sets of examples, source and target feature vectors, the traversed feature vector will more resemble the sentences in the target set. This means that if we traverse the manifold for a sentence with a different topic than the sentences in the source and target sets, the traversed vector might not preserve the topic of the original sentence.
% moved the figure from here to put in side by side with the model one
\begin{table}[h]
	\centering
	\caption[Sentences generated by the RNN]{Regenerated ($\mathbf{z}$), and traversed and generated($\mathbf{z}^*$) sentences compared to the original.
		%Sentences generated by the RNN, both from the original vector ($\mathbf{z}$) and from the traversed vector ($\mathbf{z}^*$), along with the original sentences.
	}
	\label{tab:rnn_sentences}
	\begin{tabularx}{1.0\textwidth}{l>{\footnotesize}X}%{lX}
		\hline
		Original:             & unfortunately , this is a bad movie that is just plain bad  \\
		\rowcolor{Gray}
		From $\mathbf{z}$:    & unfortunately , this is a bad movie that is just plain bad  \\
		From $\mathbf{z}^*$:  & overall , this is a good movie that is just good
		\\
		\hline
		Original:             & one of the oddest and most inexplicable sequels in movie history\\
		\rowcolor{Gray}
		From $\mathbf{z}$:    & most of the oddest and most strange movie in history history\\
		From $\mathbf{z}^*$:  & most interesting and most wonderful movie in one of the oddest ways
		\\ 
		\hline
		Original:             & still , i do like this movie for it's empowerment of women there 's not enough movies out there like this one\\
		\rowcolor{Gray}
		From $\mathbf{z}$:    & still , i do like this movie for one of adults 's not like enough like ages out there 's no women\\
		From $\mathbf{z}^*$:  & still , i do not like this movie 's not one of adults for no people who do not like this   
		\\ 
		\hline
		Original:             & i highly recommend this movie for anyone interested in art , poetry , theater , politics , or japanese history\\
		\rowcolor{Gray}
		From $\mathbf{z}$:    & i highly recommend this movie , interested for poetry , poetry , poetry , interested in history , or interested history\\
		From $\mathbf{z}^*$:  & i highly recommend this movie , except for anything , in any movie , not n't interested in any crappy movie
	\end{tabularx}
\end{table}
%------------------------------------------
\subsection{Analysis of transformed sentences}
\label{sec:eval_complete}
%------------------------------------------
There exists no single correct output for the manifold traversal, e.g given the negative sentence ``The food did not taste well'', both sentences ``The food was amazing'' and ``I liked the food'' are valid outputs that reverse the sentiment. Therefore, scores and measures used for other NLP tasks, like BLEU \citep{BLUE} for machine translation, are difficult to apply to the manifold traversal. Instead we focus on qualitative evaluation. The encoding-decoding, and the model as a whole, is evaluated by generating sentences from the feature vectors $\mathbf{z}$ (representing the original sentence) and $\mathbf{z}^*$ (the traversed vector) respectively. The generated sentences are manually compared to the original. Ideally, the sentence generated from $\mathbf{z}$ should closely resemble the original sentence while the sentence generated from $\mathbf{z}^*$ should have the same context, but opposite sentiment, as the original sentence. In table \ref{tab:rnn_sentences} some of the better examples of sentences generated by the trained RNN are shown. The overall impression is that, while having poor grammar, the model works well in terms of changing sentiment. We see that the generated sentences have the same topic as the original and that they are composed mostly by the same words. It is also found that shorter sentences are more easily encoded and decoded.
%
%Original: This movie has a cutting edge to it.
%Generated from z: <START> this movie has a edge edge to it
%Generated from z*: <START> this movie has a halt halt to suck it 

% Original: The movie I received was a great quality film for it's age.
% Generated from z: <START> the film i was a great quality quality for it 's age 
% Generated from z*: <START> the movie i was a poor movie i was given it 's a juvenile 

%Original: But besides that, the movie was terrible.
% Generated from z: <START> the bad that , the movie was horrible 
% Generated from z*: <START> the movie that , the movie was good 

%Original: I agree with Jessica, this movie is pretty bad.  
% Generated from z: <START> i agree with this movie is pretty bad , i 
% Generated from z*: <START> i agree with great good movie , this movie is great

\section{Conclusion}
An algorithm for sentiment manipulation was presented and evaluated. Visualizations of the embedding space indicate that sentence representations can be moved such that the sentiment changes while the semantics is preserved. Further, examination of generated sentences from manipulated embeddings confirmed that the sentiment had changed while the semantics and acceptability had stayed largely constant.

\subsubsection*{Acknowledgments}

The authors would like to acknowledge the project \emph{Towards a knowledge-based culturomics} supported by a framework grant from the Swedish Research Council (2012--2016; dnr 2012-5738).

\bibliography{nips_2017}
\bibliographystyle{ijcnlp2017}

\end{document}